\documentclass[conference]{IEEEtran}
\usepackage[numbers,sort&compress]{natbib}
\usepackage{graphicx}
\usepackage{dirtytalk}
\usepackage{etoolbox}
\usepackage{booktabs}
\usepackage{multirow}

\usepackage[table,xcdraw]{xcolor}
\makeatletter

\patchcmd{\@makecaption}
 {\\}
 {.\ }
 {}
 {}
 
\ifCLASSINFOpdf
\else
\fi
\usepackage{amsmath}
\usepackage{amssymb}
\usepackage{threeparttable}
\usepackage{scalerel}
\usepackage{tikz}
\usetikzlibrary{svg.path}

\definecolor{orcidlogocol}{HTML}{A6CE39}
\tikzset{
    orcidlogo/.pic={
        \fill[orcidlogocol] svg{M256,128c0,70.7-57.3,128-128,128C57.3,256,0,198.7,0,128C0,57.3,57.3,0,128,0C198.7,0,256,57.3,256,128z};
        \fill[white] svg{M86.3,186.2H70.9V79.1h15.4v48.4V186.2z}
        svg{M108.9,79.1h41.6c39.6,0,57,28.3,57,53.6c0,27.5-21.5,53.6-56.8,53.6h-41.8V79.1z M124.3,172.4h24.5c34.9,0,42.9-26.5,42.9-39.7c0-21.5-13.7-39.7-43.7-39.7h-23.7V172.4z}
        svg{M88.7,56.8c0,5.5-4.5,10.1-10.1,10.1c-5.6,0-10.1-4.6-10.1-10.1c0-5.6,4.5-10.1,10.1-10.1C84.2,46.7,88.7,51.3,88.7,56.8z};
    }
}

\newcommand\orcidicon[1]{\href{https://orcid.org/#1}{\mbox{\scalerel*{
                \begin{tikzpicture}[yscale=-1,transform shape]
                \pic{orcidlogo};
                \end{tikzpicture}
            }{|}}}}

\ifCLASSOPTIONcompsoc
\usepackage[caption=false,font=normalsize,labelfont=sf,textfont=sf]{subfig}
\else
\usepackage[caption=false,font=footnotesize]{subfig}
\fi
\usepackage{hyperref}

\begin{document}

\title{LASSR: Effective Super-Resolution Method for Plant Disease Diagnosis}

\author{\IEEEauthorblockN{Quan Huu Cap\IEEEauthorrefmark{1}, Hiroki Tani\IEEEauthorrefmark{1}, Hiroyuki Uga\IEEEauthorrefmark{2}, Satoshi Kagiwada\IEEEauthorrefmark{3}, and Hitoshi Iyatomi\IEEEauthorrefmark{1}}
\IEEEauthorblockA{huu.quan.cap.78@stu.hosei.ac.jp\quad
h.taniita@gmail.com\quad uga.hiroyuki@pref.saitama.lg.jp\\ kagiwada@hosei.ac.jp\quad iyatomi@hosei.ac.jp}
\IEEEauthorblockA{\IEEEauthorrefmark{1}Applied Informatics, Graduate School of Science and Engineering, Hosei University, Tokyo, Japan}
\IEEEauthorblockA{\IEEEauthorrefmark{2}Saitama Agricultural Technology Research Center, Saitama, Japan}
\IEEEauthorblockA{\IEEEauthorrefmark{3}Clinical Plant Science, Faculty of Bioscience and Applied Chemistry, Hosei University, Tokyo, Japan}}

\maketitle
% ABSTRACT
\begin{abstract}
    The collection of high-resolution training data is crucial in building robust plant disease diagnosis systems, since such data have a signiﬁcant impact on diagnostic performance. 
However, they are very difficult to obtain and are not always available in practice. 
Deep learning-based techniques, and particularly generative adversarial networks (GANs), can be applied to generate high-quality super-resolution images, but these methods often produce unexpected artifacts that can lower the diagnostic performance. 
In this paper, we propose a novel artifact-suppression super-resolution method that is specifically designed for diagnosing leaf disease, called Leaf Artifact-Suppression Super Resolution (LASSR). 
Thanks to its own artifact removal module that detects and suppresses artifacts to a considerable extent, LASSR can generate much more pleasing, high-quality images compared to the state-of-the-art ESRGAN model. 
Experiments based on a five-class cucumber disease (including healthy) discrimination model show that training with data generated by LASSR significantly boosts the performance on an unseen test dataset by nearly 22\% compared with the baseline, and that our approach is more than 2\% better than a model trained with images generated by ESRGAN. \\
\end{abstract}

% KEYWORDS 

% INTRODUCTION
\section{Introduction}
    Automated plant disease diagnosis is a very important part of smart agriculture. 
With the rapid development of technologies in this area, automated disease diagnosis has started to replace visual observation by experienced experts, which is prohibitively expensive and time-consuming on large farms. 

%Figure 1
\begin{figure}[!t]
\centering
\includegraphics[width=1.0\linewidth]{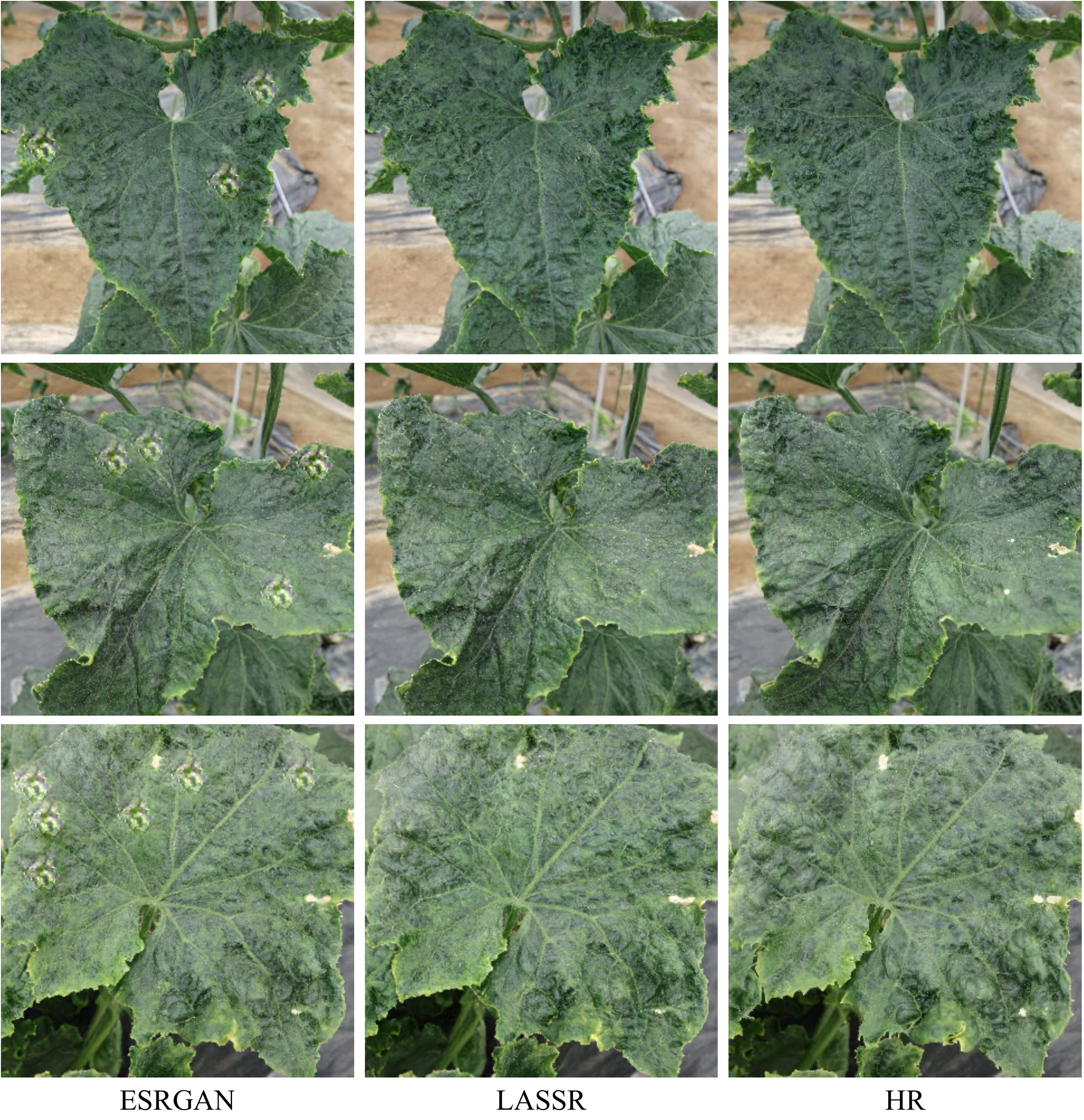}
\caption{
    Comparison of SR methods for leaf regions (4x up-sampling): (a) ESRGAN; (b) the proposed LASSR; and (c) the original HR image. The artifacts in ESRGAN can obscure the symptoms of disease and make it difficult to detect certain types of disease.
}
\label{fig:fig_1}
\end{figure}
Following the tremendous success of deep learning in the field of computer vision, numerous fast and accurate computer-based methods for plant disease diagnosis have been proposed that can effectively reduce the loss of crop yield \citep{kawasaki2015, mohanty2016, fuentes2017robust, fujita2018practical, ferentinos2018deep}. 
However, these systems require relatively close-up input images, which contain few targets for diagnosis. Applying these models to wide-angle images in large farms would be very time-consuming, since many targets (e.g. leaves) need to be diagnosed. 
Wide-angle images on practical agricultural sites contain numerous objects to be detected, many of which are visually similar and overlap each other. 
This method is therefore ineffective when applied to wide-angle images as we have experienced \citep{quan2018, suwa2019comparable}. 
Several studies have pointed out that the lack of resolution of targets for diagnosis in wide-angle images is the main reason for the relatively low diagnostic performance. 

Sa et al. \citep{sa2016deepfruits} and Bresilla et al. \citep{bresilla2019single} designed systems for the real-time counting of fruits on trees, in order to support robotic harvesting. 
However, they report that low-quality test images could cause their detection systems to miss these fruits. 
Tian et al. \citep{tian2019apple} developed an algorithm for in-farm apple fruit detection, but their scheme required images with high resolution (HR) and a high level of detail for accurate detection. 
Cap et al. \citep{quan2018} proposed an end-to-end disease diagnosis system for wide-angle images of cucumbers. 
They confirmed that small leaf sizes and low-quality input images (low-resolution, blur, poor camera focus, etc.) could significantly reduce the diagnostic performance of their disease detection scheme. 
One possible solution would be to use HR camera devices to obtain high-quality images, but this is generally expensive to deploy in practice. 

It is generally known in the main field of computer vision that a certain degree of resolution is necessary to achieve high discriminative power. 
Most ConvNets using ImageNet datasets \citep{deng2009imagenet} have long used a resolution of 224x224, following the achievements of AlexNet \citep{krizhevsky2012imagenet}, but recently, higher resolution models have been more successful. 
AmoebaNet \citep{real2019regularized} and GPipe \citep{huang2019gpipe} have achieved state-of-the-art levels of accuracy for ImageNet classification with resolutions of 331x331 and 480x480, respectively. 
Tan et al. \citep{tan2019efficientnet} proposed a scaling method called the compound coefficient that balances the depth and width (number of filters) of the network and the resolution of the input image. 
They demonstrated that their EfficientNet achieved state-of-the-art results with significantly lower computational requirements. 
As these results show, image resolution is an important factor in achieving high-accuracy recognition. 

In practical agricultural applications, we believe that recovering the high-frequency components of images by applying super-resolution (SR) methods offers a promising solution for addressing the abovementioned issue. 
SR techniques can be divided into two broad types called \say{registration-type} and \say{learning-type}. 
Registration-type SR techniques utilize a large number of images in order to increase the pixel density of the image. 
In this way, the true high-frequency components of the image can be estimated using an appropriate reconstruction algorithm. 
Typical registration-type SR techniques utilize maximum likelihood (ML) \citep{tom1995reconstruction}, maximum a posteriori (MAP) methods \citep{schultz1996extraction}, or projection onto convex sets (POCS) \citep{patti2001artifact} as a reconstruction algorithm. 
However, since these classical methods require a relatively high number of observed images, precise correction of the positional deviations between images using sub-pixel image registration is necessary for the successive reconstruction process. 
SR methods using multi-camera devices \citep{hirao2015prototype,quevedo2017underwater} also fall into this category; although these techniques have been commercialized in recent years with the spread of high-end mobile phone cameras, they were originally expensive. 

Learning-type SR techniques usually utilize only one base image from the observed set, and predict unknown details. 
Before deep learning techniques were developed, they often applied pre-trained database and/or estimators \citep{freeman2002example} or an interpolation approach based on signal processing techniques \citep{komatsu2010super}. 
The quality and resolution of SR images generated by these methods were usually lower than the other and required appropriate settings. 
However, thanks to the modeling power of convolutional neural networks (CNNs), recent SR methods based on a single image, known as single image super-resolution (SISR) techniques, have shown excellent performance \citep{dong2015image, ledig2017photo, wang2018esrgan}. 
Dong et al. \citep{dong2015image} first proposed the super-resolution convolutional neural network (SRCNN), which provided end-to-end training between low-resolution (LR) and HR images. 
Ledig et al. \citep{ledig2017photo} then proposed SRGAN as the first SR method to adopt a generative adversarial network (GAN) \citep{goodfellow2014generative} algorithm, resulting indistinguishable super-resolved images from high-resolution images. 
Recently, an improved version of SRGAN called ESRGAN \citep{wang2018esrgan} with the proposed residual-in-residual dense block (RRDB) and the relativistic average GAN (RaGAN) \citep{jolicoeur2018relativistic} loss, outperformed SRGAN in terms of perceptual quality. 

SR techniques have been widely used in various fields, however, only limited applications in the agricultural sector have been reported thus far \citep{kasturiwala2015adaptive,yamamoto2017super,cap2019super,dai2020crop}. 
Kasturiwala et al. \citep{kasturiwala2015adaptive} applied an iterative curvature-based interpolation method \citep{giachetti2011real} to increase the resolution of diseased leaf images. 
They claimed this approach could support pathologists with better visual quality of the infected leaves, but have not yet tested their method on any disease diagnosis tasks. 
Yamamoto et al. \citep{yamamoto2017super} and Dai et al. \citep{dai2020crop} improved disease diagnostic performance by applying an SRCNN and a GAN-based SR model called DATFGAN to tomatoes and other types of crops, respectively. 

Although these methods showed promising results, they are not very realistic, since they were applied to the impractical PlantVillage dataset \citep{hughes2015open} in which each leaf image was taken under ideal conditions (e.g. manually cropped and placed on a uniform background). 
Several reports have shown that the diagnostic performance of systems trained on these images is significantly reduced when applied to real on-site images \citep{mohanty2016,ferentinos2018deep}. 
Hence, we cannot conclude from their results that their diagnostic schemes can be used in practice, or that SR contributes to improving diagnostic accuracy in practical situations. 

To confirm the effectiveness of SR in practical situations, we proposed a GAN-based SR method for the in-field diagnosis of cucumber disease \citep{cap2019super}. l
We found that the use of an SR model with a perceptual loss function \citep{johnson2016perceptual} dramatically improved the diagnostic performance. 
The diagnostic results from our SR model were 20.7\% better than those of the state-of-the-art SRResNet model \citep{ledig2017photo} at the time. 

However, we experienced that SR images generated using GAN often contain artifacts like \say{rubber stamps}, especially in the leaf region (see Fig. \ref{fig:fig_1}). 
Leaves have been most frequently targeted in the study of automatic diagnosis of plants, since they exhibit more disease characteristics than other parts of the plant. 
We are particularly mindful of the problem of artifacts occurring in the leaf region, as this could cause difficulty in diagnosing some types of disease in practical situations. 
To address this problem, we propose an effective artifact-suppression SR method specifically designed for leaves, called Leaf Artifact-Suppression Super Resolution (LASSR). 
%Figure 2
\begin{figure*}[!t]
\centering
\includegraphics[width=0.75\linewidth]{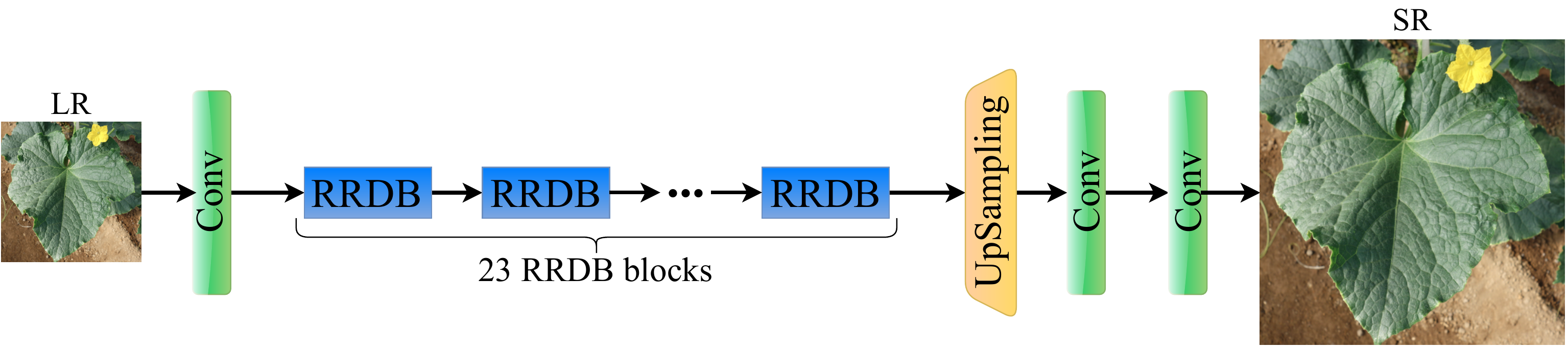}
\caption{
    The generator $G$ consists of 23 RRDB blocks, followed by $4\times$ upsampling and convolutional layers to generate the SR images.
}
\label{fig:fig_2}
\end{figure*}
%Figure 3
\begin{figure*}[!t]
\centering
\includegraphics[width=0.76\linewidth]{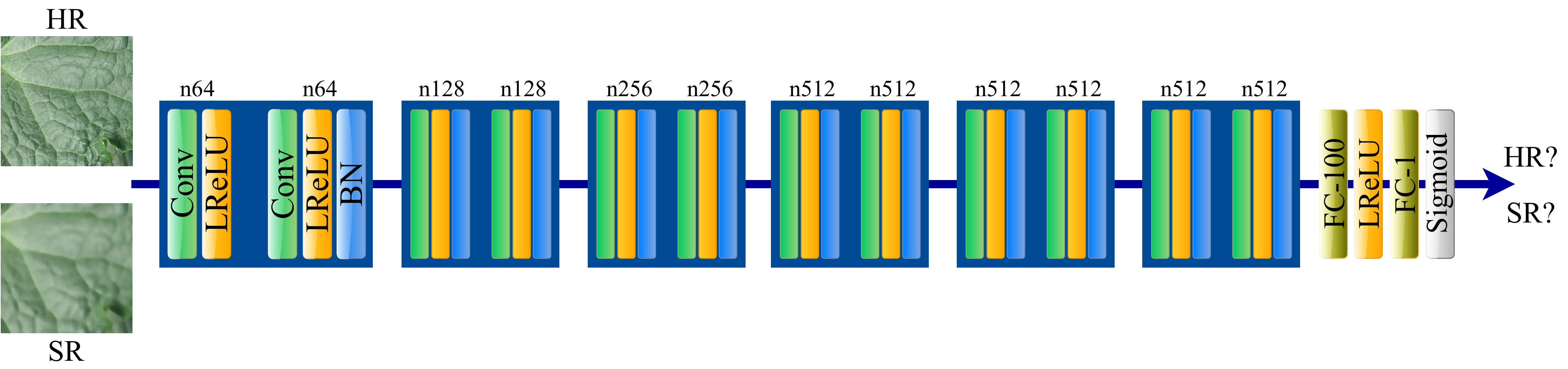}
\caption{
    The discriminator $D$, consisting of six \emph{conv\_blocks} with corresponding number of feature maps ($n$).
}
\label{fig:fig_3}
\end{figure*}

One further aspect of this proposal should be emphasized. 
In the field of automated plant diagnosis, evaluation data are in most cases generated from a portion of the training dataset. 
Recently, it has been noted that the reported diagnostic accuracy is likely to have been superficially overstated due to \say{latent similarities} within the dataset (i.e., similarities in image conditions such as background and light conditions) \citep{quan2018,suwa2019comparable,saikawa2019aop}. 
We therefore evaluate the performance of our LASSR method based on the degree of improvement in diagnostic accuracy for a completely exclusive dataset, in addition to the image quality. 

In this study, we found that the larger the size of the image input to the model, the better the diagnostic accuracy achieved on an \emph{unseen dataset}. 
However, HR training resources are not always available in practice. 
We therefore believe that SR methods can be used to generate high-quality training resources and can help to improve the robustness of disease diagnosis systems on unknown test data to allow for more practical use. 

In summary, the contributions of this work are as follows: 
\begin{itemize}
    \item We propose LASSR as a specially designed SR method to improve the performance of plant leaf disease diagnosis, with a novel artifact removal module (ARM) that dynamically suppresses artifacts on-the-fly during training. 
    \item LASSR provides visually pleasing images by effectively suppressing artifacts and gives a better Fréchet inception distance (FID) \citep{heusel2017gans} than the ESRGAN method. 
    \item LASSR significantly improves the accuracy of plant diagnosis in unseen images by over 22\% from the baseline. This is more than 2\% better than a model trained on images generated by ESRGAN. 
\end{itemize}

\section{Proposed Method - LASSR}
    The proposed LASSR is a SISR framework that is specifically designed to solve the problem of artifacts in SR images and hence to improve the performance of plant disease diagnosis. 
LASSR is basically built on ESRGAN \citep{wang2018esrgan} with the proposed artifact removal module (ARM) to guide the network in suppressing artifacts on-the-fly during training. 
Fig. \ref{fig:fig_1} shows examples of the artifacts generated by the ESRGAN model; our proposed LASSR resolves this problem, resulting in natural and convincing generated images. %Figure 4
\begin{figure*}[!t]
\centering
\includegraphics[width=0.7\linewidth]{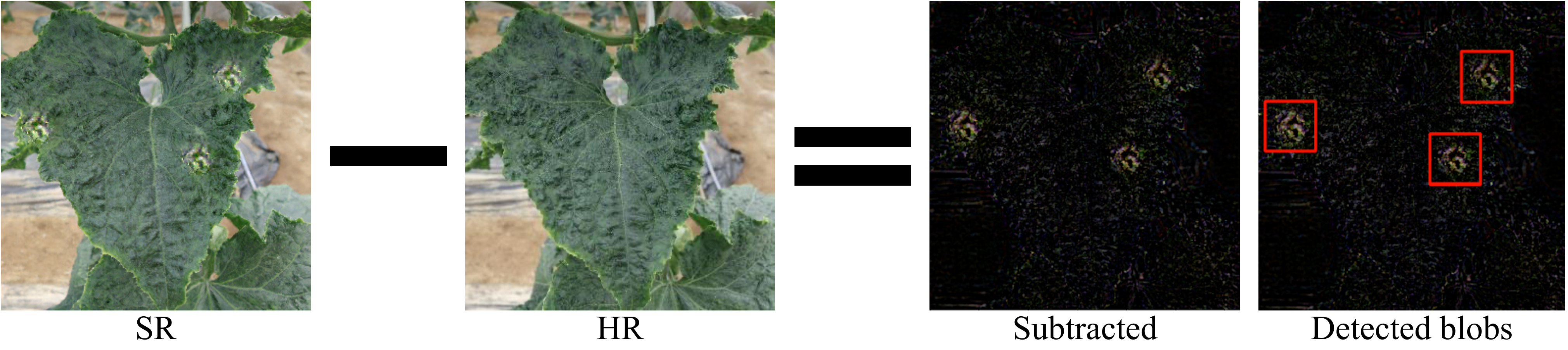}
\caption{
    SR and ground-truth HR images of a leaf. After obtaining the subtracted image, the blobs can be detected by applying DoG.
}
\label{fig:fig_4}
\end{figure*}
\subsection{Network Architectures}
LASSR is inherited from our previous GAN-based network \citep{cap2019super}. 
It is composed of two CNN models: the generator $G$, which generates SR images, and the discriminator $D$, which distinguishes SR images from HR images. 
The networks are trained together to solve an adversarial min-max problem. 

\subsubsection{The Generator}
Our scheme uses the architecture of a generator $G$, in the same way as in ESRGAN \citep{wang2018esrgan}. 
$G$ is composed of 23 residual-in-residual dense (RRDB) blocks, resulting in a total of 115 convolutional layers. 
Our network $G$ up-scales 4x from the input LR image. 
Fig. \ref{fig:fig_2} illustrates the architecture of the generator $G$ used in our experiments. 
The reader is referred to the ESRGAN paper for more technical details. 

\subsubsection{The Discriminator}
Our discriminator $D$ is designed in the same way as in our previous model \citep{cap2019super}. 
It is deeper than the discriminator used in ESRGAN, and has a larger input size of 192x192. 
The architecture of our discriminator $D$ is illustrated in Fig. \ref{fig:fig_3}. 
We define a convolution block as a block of two convolutional (Conv) layers. 
Following a Conv, we use either a leaky rectified linear function (LReLU) \citep{maas2013rectifier} or a combination of LReLU and batch normalization (BN) \citep{ioffe2015batch}. 
We use LReLU with $\alpha=0.2$ as the activation function for all layers except for the last. The reader is referred to our previous paper \citep{cap2019super} for more details. 

\subsection{Loss Functions of LASSR}
The objective functions in LASSR are extended from ESRGAN. 
To train our generator $G$, we minimize the loss function $\mathcal{L}_\mathrm{G}$ as follows:
%% Equation (1)
\begin{equation}
\mathcal{L}_\mathrm{G}=\lambda \mathcal{L}_\mathrm{G}^\mathrm{adv}+\mathcal{L}_\mathrm{percep}+\eta{\left|I_\mathrm{HR}-I_\mathrm{SR}\right|}_1+\beta \mathcal{L}_\mathrm{ARM},
\end{equation}
where $\mathcal{L}_\mathrm{G}^\mathrm{adv}$, $\mathcal{L}_\mathrm{percep}$, and ${\left|I_\mathrm{HR}-I_\mathrm{SR}\right|}_1$ appear in the original loss function in ESRGAN. 
Here, $\mathcal{L}_\mathrm{G}^\mathrm{adv}$  is the adversarial loss for the generator $G$, and $\mathcal{L}_\mathrm{percep}$ is the perceptual loss \citep{johnson2016perceptual}, which minimizes the similarity between the HR and SR images in the feature space of the VGG19 model \citep{Simonyan15} pre-trained with the ImageNet dataset \citep{deng2009imagenet}. 
$I_\mathrm{HR}$ and $I_\mathrm{SR}$ are the HR and SR images, respectively. 
$\mathcal{L}_\mathrm{ARM}$ is our proposed novel loss term for calibrating the artifact effects, and is formed based on our proposed ARM (described in detail in the next section). 
$\lambda$, $\eta$, and $\beta$ are coefficients used to balance the different loss terms. 

To train our discriminator $D$, we use the same adversarial loss $\mathcal{L}_\mathrm{D}$ as in ESRGAN: 
%% Equation (2)
\begin{equation}
\mathcal{L}_\mathrm{D}=-\mathbb{E}_{I_\mathrm{HR}}[\mathrm{log}(D(I_\mathrm{HR},I_\mathrm{SR}))]-\\
\mathbb{E}_{I_\mathrm{SR}}[\mathrm{log}(1-D(I_\mathrm{SR},I_\mathrm{HR}))].
\end{equation}
Finally, the $G$ and $D$ networks are trained together to solve an adversarial min-max problem \citep{goodfellow2014generative}. 
For more details of the loss functions, the reader is referred to the ESRGAN literature \citep{wang2018esrgan}. 

\subsubsection{The Artifact Removal Module}
In this work, we propose a novel ARM that detects and suppress artifacts on-the-fly during the training of our SR model. 
The ARM acts like a dynamic training strategy, and provides guidance allowing LASSR to suppress the occurrence of artifacts. 
The key idea of the ARM is to detect artifacts and then to minimize the differences between the areas of these artifacts and the corresponding areas of the ground-truth HR images. 
In most cases, artifacts appear in the form of similar specks or rubber stamps, as shown in Fig. \ref{fig:fig_1}. 
We refer to these artifact regions as blobs, and apply the difference of Gaussians (DoG) to detect them. 
DoG is an algorithm that finds scale-space maxima by subtracting different blurred versions of an original image. 
These blurred images are obtained by convolving the original images with Gaussian kernels with differing standard deviations. 
After obtaining the scale-space maxima, the areas of blobs are defined by the local maxima points and their corresponding Gaussian kernels. 

Fig. \ref{fig:fig_4} illustrates the steps in the process of detecting artifact areas (blobs) in SR images. 
Given a SR image and a ground-truth HR image, we first subtract two images to obtain the subtracted version. 
DoG is then applied to detect the locations of artifacts (blobs). 
Based on our preliminary experiments, artifact areas can be detected effectively using only two Gaussian kernels with corresponding values of $\sigma_1=0.078 \times N$ and $\sigma_2=0.104 \times N$, where $N \times N$ is the size of the cropped image used for training (i.e. $192 \times 192$ in our study). 

Once the artifacts have been detected, we add a new loss term $\mathcal{L}_\mathrm{ARM}$ in Eq. (1) to LASSR to allow the network to suppress the artifact. $\mathcal{L}_\mathrm{ARM}$ is defined as:
%% Equation (2)
\begin{equation}
\mathcal{L}_\mathrm{ARM}=\sum_{b_i\in\mathfrak B}b_i,
\end{equation}
where $b_i$ is the sum of the pixel values of a blob detected in the subtracted image, and $\mathfrak B=\left\{b_1,b_2,...,b_{\left|\mathfrak B\right|}\right\}$ is the set of detected blobs. 
% Table I
\begin{table}[!t]
\centering
\caption{Statistics for Dataset-B$\dagger$}
\label{tab:table-I}
\begin{threeparttable}
\begin{tabular}{|c|c|c|c|}
\hline
\textbf{Class} & Dataset-$\mathrm{B}_\mathrm{Train}$ & Dataset-$\mathrm{B}_\mathrm{Val}$ & Dataset-$\mathrm{B}_\mathrm{Test}$ \\ \hline
Healthy        & 13,089             & 4,394              & 1,276              \\ \hline
Brown spot     & 5,142              & 1,668              & 2,786              \\ \hline
CCYV           & 4,356              & 1,438              & 2,096              \\ \hline
MYSV           & 10,451             & 3,512              & 1,550              \\ \hline
Downy mildew   & 2,514              & 893                & 2,219              \\ \hline
Total          & 35,552             & 11,905             & 9,927              \\ \hline
\end{tabular}
\begin{tablenotes}
    \item[$\dagger$] {Dataset used to evaluate the performance of the SR schemes}
\end{tablenotes}
\end{threeparttable}
\end{table}

\section{Experiments}
    We conducted two experiments: the first evaluated the quality of SR images, while the second evaluated the improvement in disease diagnostic performance achieved by using training images obtained with SR techniques. 
This evaluation was used due to the difficulty in obtaining HR images, which usually provide high diagnostic accuracy, as mentioned above.
%Figure 5
\begin{figure*}[!t]
\centering
\includegraphics[width=0.9\linewidth]{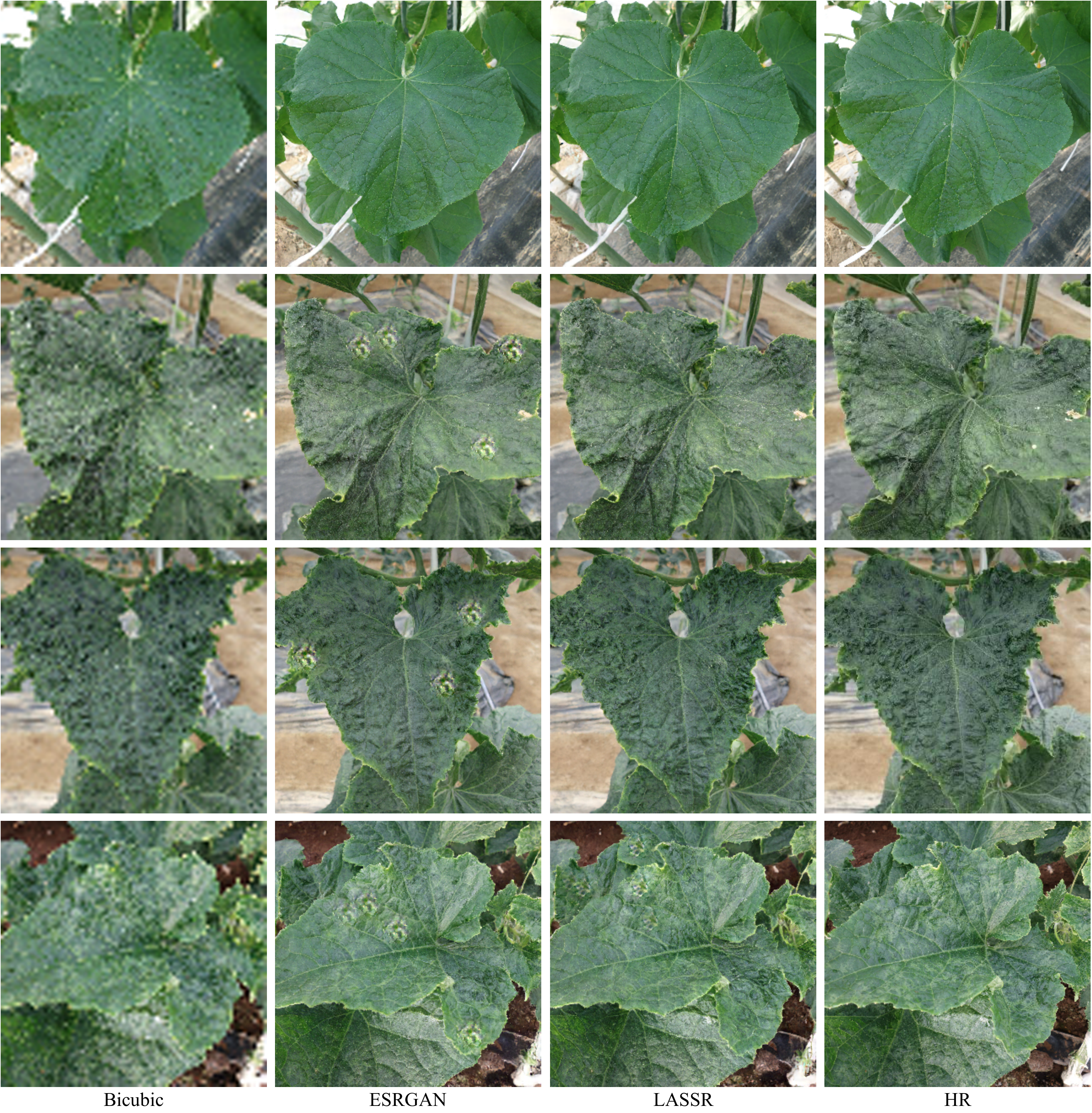}
\caption{
    Comparison between the generated SR results and the original HR images. LASSR generates more a natural image with very suppressed artifacts compared to the existing ESRGAN method.
}
\label{fig:fig_5}
\end{figure*}
\subsection{Datasets}
In these experiments, we used two datasets, Dataset-A and Dataset-B. 
Dataset-A was used to train and evaluate both LASSR and the ESRGAN model as a comparison. 
Dataset-B was used to train and evaluate the classifiers, in order to assess the effectiveness of the disease diagnostic ability of SR. 
These datasets are independent of each other. 

\subsubsection{Dataset-A for SR Models}
In this experiment, we used a cucumber dataset previously reported in the literature \citep{hiroki2018diagnosis,cap2019super} as Dataset-A. 
This is a multiple infection dataset with 25 classes containing a total of 48,311 cucumber leaf images, of which 38,821 show single infections, 1,814 show multiple infections, and 7,676 contain healthy leaves. 
Each image has a size of $316\times 316$ pixels. 
We divided this dataset into separate training and testing datasets. 
The training set contained 36,233 images (roughly 75\% of the dataset, referred to here as Dataset-$\mathrm{A}_\mathrm{Train}$), while the testing set contained 12,078 images (roughly 25\% of the dataset, referred to as Dataset-$\mathrm{A}_\mathrm{Test}$).

\subsubsection{Dataset-B for Disease Classifier}
Dataset-B was another cucumber leaf dataset collected from multiple locations in Japan, taken during the period 2015–2019. 
Table \ref{tab:table-I} summarizes the statistics for this dataset. 
It contained four classes of disease (\emph{Cucurbit chlorotic yellows virus} (CCYV), \emph{Melon yellow spot virus} (MYSV), \emph{Brown spot}, \emph{Downy mildew}) and healthy. 
Each image in this dataset had a size of $512 \times 512$ pixels. 
We divided this dataset into two sets, Dataset-$\mathrm{B}_\mathrm{Train/Val}$ and Dataset-$\mathrm{B}_\mathrm{Test}$. 
Images in Dataset-$\mathrm{B}_\mathrm{Test}$ were taken at different times and in different locations from those in Dataset-$\mathrm{B}_\mathrm{Train/Val}$ in order to avoid the problem of latent similarities among datasets, as mentioned earlier. 
Note that the appearances of the images in these sets varied widely, due to the differences in the circumstances (i.e. photographic conditions and background) in which they were taken.

\subsection{Experiment I: LASSR Improving Image Quality}
We first evaluated the improvement in image quality by LASSR and compared it to existing ESRGAN methods. 
We used FID scores \citep{heusel2017gans} as our evaluation criteria, in the same way as in other SR methods.  
This is because other standard quantitative measures such as PSNR and SSIM have been reported as being unable to capture and accurately assess the perceptual image quality which is highly correlated with human perception \citep{toderici2017full,ledig2017photo,zhang2018unreasonable}. 
We trained our LASSR and ESRGAN using Dataset-$\mathrm{A}_\mathrm{Train}$. 
During training, HR images were obtained by random cropping from training images with the size of $192 \times 192$. 
LR images were then created by $1/4 \times$ down-sampling from HR images, using bicubic interpolation. 
Both LR and HR images were augmented with random horizontal flips and random 90 degrees rotations on-the-fly. 
It should be noted that these are the same conditions that were used in the original training of ESRGAN. 

The ESRGAN model was trained using the loss functions from the original paper, while our LASSR was trained using the loss functions in Eqs. (1) and (2), with $\lambda = \beta = 5 \times 10^{-3}$ and $\eta = 10^{-2}$. We set the mini-batch size to 32 images and used the Adam optimizer \citep{kingma2015adam} for both $G$ and $D$ models. The training process was completed after 400 epochs. More details are given in the Results section.
%Figure 6
\begin{figure}[!t]
\centering
\includegraphics[width=0.99\linewidth]{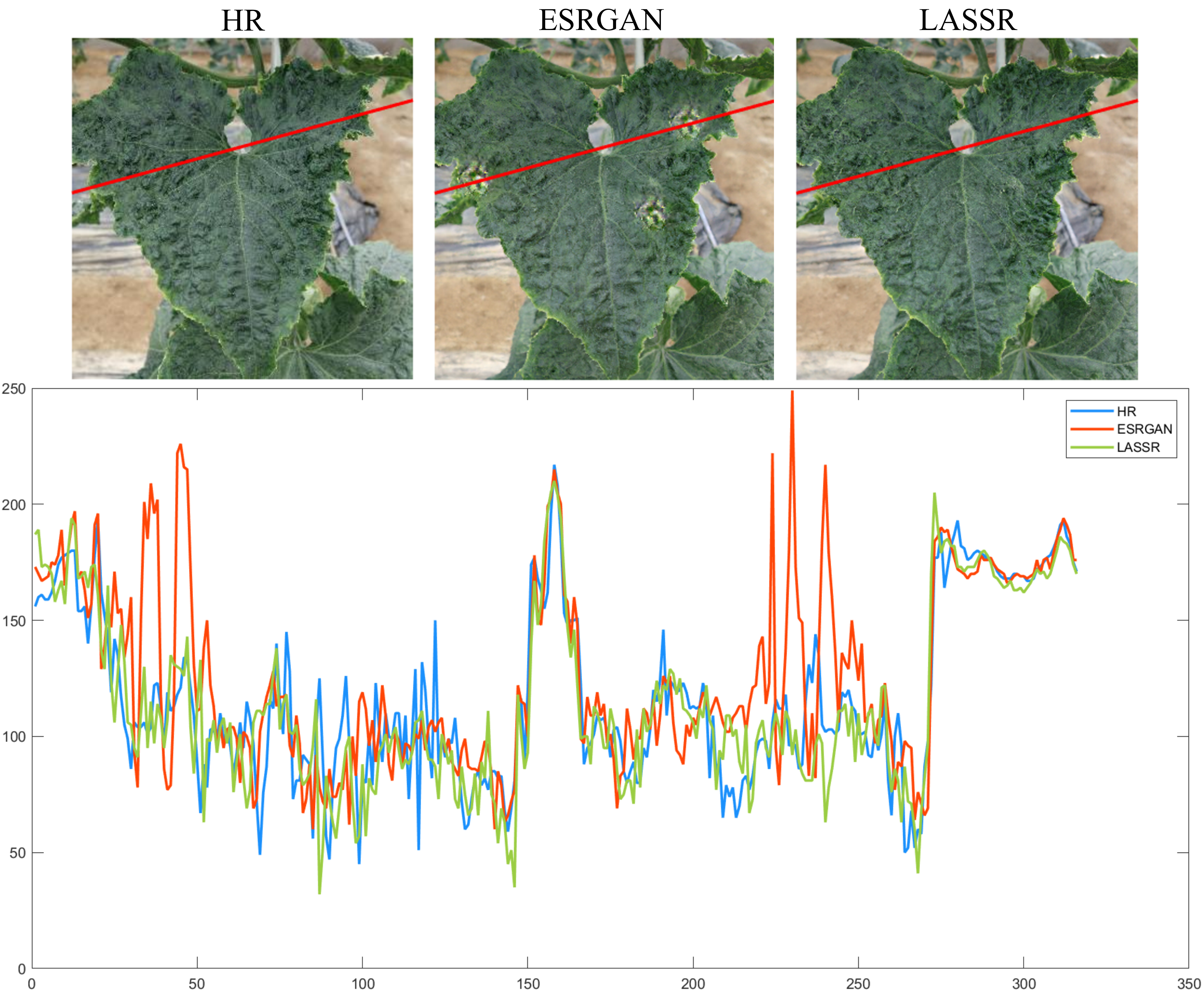}
\caption{
    Line profiles for the SR and the original HR images. We can see similar profiles of LASSR and HR, while that of ESRGAN shows significant dissimilarities due to the presence of artifacts.
}
\label{fig:fig_6}
\end{figure}
\subsection{Experiment II: LASSR Improving Diagnostic Performance on an Unseen Dataset}
In this experiment, we trained similar plant disease classifiers using LR, HR and SR images using both LASSR and ESRGAN for comparison. 
Specifically, we trained the following five classiﬁers based on the pre-trained ResNet50 \citep{he2016deep} model: 
\renewcommand{\theenumi}{\arabic{enumi}}
\begin{enumerate}
    \item ResNet\_LR: The classifier was trained on a 1/4 x down-sampled Dataset-$\mathrm{B}_\mathrm{Train}$ (i.e. input size $128 \times 128$) using bicubic interpolation.
    \item ResNet\_HR: The classifier was trained with Dataset-$\mathrm{B}_\mathrm{Train}$ (i.e. input size $512 \times 512$).
    \item ResNet\_Bicubic: As in (1), except with training images 4x SRed (i.e. $512 \times 512$) using bi-cubic interpolation.
    \item ResNet\_ESRGAN: As in (1), except with training images 4x SRed (i.e. $512 \times 512$) using ESRGAN.
    \item ResNet\_LASSR [proposed model]: As in (1), except with training images 4x SRed (i.e. $512 \times 512$) using LASSR.
\end{enumerate}
All five ResNet50-based classifiers were fine-tuned at all layers with the Adam optimizer \citep{kingma2015adam}. 
To handle the class imbalance in Dataset-$\mathrm{B}_\mathrm{Train}$, we applied the softmax class-balanced loss \citep{cui2019class} with $\beta=0.9999$ to all classifiers. 
During the training process, we applied random horizontal and vertical flips to each image. 
Training was complete after 20 epochs. 

\section{Results}
    \subsection{LASSR: Improving Image Quality}
Figs. \ref{fig:fig_5} and \ref{fig:fig_6} show the visual comparison and line profiles of the generated images and the original HR image.
Table \ref{tab:table-II} shows the FID scores of the images generated using the bicubic, ESRGAN and LASSR techniques versus the original HR images. 
% Table II
\begin{table}[!t]
\centering
\caption{FID scores for bicubic, ESRGAN, LASSR, and HR schemes (lower is better)}
\label{tab:table-II}
\begin{threeparttable}
\begin{tabular}{|c|c|c|c|}
\hline
\textbf{Dataset} & (Bicubic, HR) & (ESRGAN, HR) & (LASSR, HR)   \\ \hline
Dataset-$\mathrm{A}_\mathrm{Test}^\ast$        & 104.49        & 2.98         & \textbf{2.90} \\ \hline
Dataset-$\mathrm{B}^{\ast\ast}$        & 45.36         & 2.42         & \textbf{2.38} \\ \hline
\end{tabular}

\begin{tablenotes}
    \item[$\ast,\ast\ast:$] {Calculated for images of size 316x316 and 512x512, respectively}
\end{tablenotes}
\end{threeparttable}
\end{table}
In a similar way as for Dataset-$\mathrm{A}_\mathrm{Test}$, we calculated the FID score for the entire Dataset-B, since it is exclusive from Dataset-$\mathrm{A}_\mathrm{Train}$. Note that the test images were $1/4 \times$ down-sampled before being fed to the SR models. 

Our LASSR scheme successfully suppressed the artifacts, and generated images that were more natural than the ESRGAN method. 
We also observed that for Dataset-$\mathrm{A}_\mathrm{Test}$, ESRGAN produced over 1,300 artifact images (11.39\% of the dataset), while our LASSR created only 177 cases (1.47\%). 
On Dataset-B, which was exclusive of Dataset-A, ESRGAN still produced 380 cases of artifacts (0.66\%) while our LASSR generated zero artifact (artifact-free) images. 
LASSR also achieved a better (lower) FID score than ESRGAN on both datasets. 

\subsection{LASSR: Improving Diagnostic Performance on Unseen Dataset}
Table \ref{tab:table-III} presents a comparison of the diagnostic performance of disease classiﬁers in Experiment II. 
We can see that there is a large gap between the micro-average accuracy of the Dataset-$\mathrm{B}_\mathrm{Val}$ and Dataset-$\mathrm{B}_\mathrm{Test}$, since the two sets are inherently different in nature due to differences in the photographed locations of the images and other factors as mentioned earlier. 
However, our LASSR model significantly helped boosting the performance on unseen dataset from the baseline model, achieving a competitive result. 

\section{Discussion}
    We proposed a SR method for addressing the artifact problem and explored its potential for improving the performance of an automated plant disease diagnosis system on unseen data.

\subsection{Improvement in Image Quality}
From the results of the Experiment I, thanks to the introduction of the novel ARM, our LASSR effectively addresses the problem of artifacts and produces more natural SR images compared to the results from ESRGAN (see Figs. \ref{fig:fig_1}, \ref{fig:fig_5} and \ref{fig:fig_6}), and achieves better FID scores (Table \ref{tab:table-II}). 
Although some artifact cases remain in the images generated by LASSR on Dataset-$\mathrm{A}_\mathrm{Test}$, it should be noted that our LASSR produced much weaker artifacts than ESRGAN (see Fig. \ref{fig:fig_5}, last row). 
Moreover, LASSR generated artifact-free SR images on Dataset-B.

The incidence of ESRGAN artifacts was 11.4\% for Dataset-$\mathrm{A}_\mathrm{Test}$, which consisted of images taken in the same field as the training data. 
The rate was far lower for Dataset-B at 0.66\%, on the completely unknown dataset. 
In addition, the artifacts generated by ESRGAN were almost all of the same size (where each side of the boundary box was about 1/10th the size of the image) and had circle-like shapes, as if the image had been stamped with a rubber stamp. 
This may be due to the particular combination of the input image and kernels in a relatively forward convolutional layer corresponding to a specific receptive field size. 
However, we have not been able to identify the reason for this. 
Although our ARM effectively suppressed and reduced these artifact effects, it seems likely that further investigation of the behavior of the convolutional layers could be a way to be toward the artifact-free SR, and we aim to address this in the near future.
% Table III
\begin{table}[!t]
\centering
\caption{Summary of results on Experiment II}
\label{tab:table-III}
\begin{tablenotes}
    \centering
    \item[] {a) Classification performance on Dataset-B (in micro-average)}
\end{tablenotes}
\begin{tabular}{|l|c|c|c|}
\hline
\multicolumn{1}{|c|}{\textbf{Model}} & Dataset-$\mathrm{B}_\mathrm{Train}$      & Dataset-$\mathrm{B}_\mathrm{Val}$      & Dataset-$\mathrm{B}_\mathrm{Test}$      \\ \hline
ResNet\_LR                           & \textbf{99.45} & 87.25          & 54.64          \\ \hline
ResNet\_Bicubic                      & 93.43          & 89.34          & 61.12          \\ \hline
ResNet\_ESRGAN                       & 94.72          & 92.66          & 74.98          \\ \hline
ResNet\_LASSR                        & 96.29          & 95.20          & 77.20          \\ \hline
ResNet\_HR                           & 96.22          & \textbf{97.59} & \textbf{82.66} \\ \hline
\end{tabular}

% Second sub table
\begin{tablenotes}
    \centering
    \item[] {b) Classification performance on Dataset-$\mathrm{B}_\mathrm{Test}$}
\end{tablenotes}
\centering
\resizebox{\linewidth}{!}{
\begin{threeparttable}
\begin{tabular}{|l|c|c|c|c|c|c|}
\hline
\multicolumn{1}{|c|}{\textbf{Model}} & \textbf{Healthy} & \textbf{Brown spot} & \textbf{CCYV}  & \textbf{MYSV} & \textbf{\begin{tabular}[c]{@{}c@{}}Downy\\ mildew\end{tabular}} & \textbf{\begin{tabular}[c]{@{}c@{}}Macro-\\ average\end{tabular}} \\ \hline
ResNet\_LR                           & 85.50   & 71.21               & 37.45          & 60.65         & 28.12                                                           & 56.59                                                             \\ \hline
ResNet\_Bicubic                      & 86.21            & 74.01               & 23.09          & 63.42         & 64.80                                                           & 62.31                                                             \\ \hline
ResNet\_ESRGAN                       & \textcolor{red}{\textbf{90.20}}            & \textcolor{blue}{\textbf{91.35}}               & 59.21          & 71.61         & 62.91                                                           & 75.06                                                             \\ \hline
ResNet\_LASSR                        & 86.60            & \textcolor{red}{\textbf{91.71}}               & \textcolor{blue}{\textbf{63.88}}          & \textcolor{blue}{\textbf{77.55}}         & \textcolor{blue}{\textbf{65.93}}                                                           & \textcolor{blue}{\textbf{77.13}}                                                             \\ \hline
ResNet\_HR                           & \textcolor{blue}{\textbf{89.11}}            & 90.42      & \textcolor{red}{\textbf{74.38}} & \textcolor{red}{\textbf{79.55}}         & \textcolor{red}{\textbf{79.22}}                                                           & \textcolor{red}{\textbf{82.54}}                                                             \\ \hline
\end{tabular}
\begin{tablenotes}
    \item[] {\textcolor{red}{\textbf{Red}} indicates the best performance and \textcolor{blue}{\textbf{blue}} indicates the second-best performance}
\end{tablenotes}

\end{threeparttable}
}
\end{table}
\subsection{Improvement in Disease Diagnosis Performance}
In Experiment II, although the diagnostic accuracy of all of the classifiers on the two subsets Dataset-$\mathrm{B}_\mathrm{Train/Val}$ was similar and very high, the performance was significantly reduced compared to Dataset-$\mathrm{B}_\mathrm{Test}$. 
As discussed above, this is due to the problem of latent similarities between data \citep{quan2018,suwa2019comparable,saikawa2019aop}. 
As frequently reported in the literature, the performance was higher for data from the same imaging environment (Dataset-$\mathrm{B}_\mathrm{Train/Val}$) due to overfitting, but lower for data from a different environment (Dataset-$\mathrm{B}_\mathrm{Test}$). 
This reduction in performance is generally known as a covariate shift. 

The ResNet\_LR and ResNet\_Bicubic models showed poor performance on Dataset-$\mathrm{B}_\mathrm{Test}$ (with micro-average accuracies of only around 54\% and 61\%, respectively, as shown in Table \ref{tab:table-III}.a). 
This was because the size of the LR input images was not sufficient for the diagnosis of plant diseases. 
Our LASSR scheme successfully reconstructed the information from LR images and helped the ResNet\_LASSR model to achieve high diagnostic performance, significantly outperforming the model trained on LR images (ResNet\_LR) by nearly 22\% in terms of micro-average accuracy (Table \ref{tab:table-III}.a). 
Moreover, ResNet\_LASSR achieved performance that was 2.2\% and 2.1\% better than ResNet\_ESRGAN in terms of micro- and macro-average accuracy, respectively (Table \ref{tab:table-III}). 

For the healthy plants, ResNet\_LASSR (86.6\%) was less numerically accurate than ResNet\_ESRGAN (90.2\%), unlike in the cases of disease. 
In this case, ResNet\_LR also shows a high degree of accuracy in diagnosis (85.5\%), and the difference from ResNet\_HR (89.1\%) is small, with ResNet\_LASSR falling in between. 
This may be due to the fact that healthy cases have essentially no signs of disease that can be detected, and there is therefore little need for HR information. 
Although we cannot determine a direct reason for the inferiority of ResNet\_LASSR compared to ResNet\_ESRGAN, we believe that the diagnostic ability of ResNet\_LASSR for healthy plants is sufficiently effective from a practical standpoint. 

It should also be noted that in terms of CCYV diagnosis, there was a reversal in the performance of ResNet\_LR and ResNet\_Bicubic. 
It appears that neither model could detect the essential features of this disease, although the latter had a wastefully larger degree of freedom, which reduced its generalizability. 

Except for the healthy images, ResNet\_LASSR performed significantly better than ResNet\_ESRGAN in all cases, achieving the closest result to the model using the original HR images. 
This reinforces our argument for the benefits of using the SR method. Since high-quality training data are not always available in practice, SR techniques are confirmed to be effective in generating reliable training resources and improving the robustness of diagnosis systems. 
Note that we also trained another classifier with HR images of size $224\times 224$ (as commonly used in many disease classifiers) and recorded a diagnostic of around 72\% in terms of micro-accuracy. This implies that the quality of the training image is important. 

Although LASSR achieved promising results, we note that the selection of hyper-parameters for the ARM is still done manually, as it depends on the input size. 
We believe that the development of an ARM that dynamically adapts the training set will further improve the effectiveness of LASSR. 

\section{Conclusion}
    We have proposed an artifact-suppression SR method called LASSR (Leaf Artifact-Suppression Super Resolution), which is specifically designed for the automatic diagnosis of plant disease. 
Our LASSR model with the novel ARM effectively addresses the artifact effects produced by a GAN-based network and improves the performance of automated plant leaf disease diagnosis. 
The proposed LASSR scheme is capable of generating high-quality images and significantly improves disease diagnostic performance on unknown images in practical settings. 
From this perspective, we have confirmed that LASSR can be used as an efficient and reliable SR tool for use in real cultivation scenarios. 
Further research on the application of LASSR to other food crops is currently being carried out.

\section*{Acknowledgment}
This research was conducted as part of a research project on future agricultural production and was supported by the Ministry of Agriculture, Forestry and Fisheries of Japan.

\nocite{*}
\footnotesize{
\bibliographystyle{IEEEtran}
\bibliography{reference}

\begin{thebibliography}{10}
\providecommand{\url}[1]{#1}
\csname url@samestyle\endcsname
\providecommand{\newblock}{\relax}
\providecommand{\bibinfo}[2]{#2}
\providecommand{\BIBentrySTDinterwordspacing}{\spaceskip=0pt\relax}
\providecommand{\BIBentryALTinterwordstretchfactor}{4}
\providecommand{\BIBentryALTinterwordspacing}{\spaceskip=\fontdimen2\font plus
\BIBentryALTinterwordstretchfactor\fontdimen3\font minus
  \fontdimen4\font\relax}
\providecommand{\BIBforeignlanguage}[2]{{%
\expandafter\ifx\csname l@#1\endcsname\relax
\typeout{** WARNING: IEEEtran.bst: No hyphenation pattern has been}%
\typeout{** loaded for the language `#1'. Using the pattern for}%
\typeout{** the default language instead.}%
\else
\language=\csname l@#1\endcsname
\fi
#2}}
\providecommand{\BIBdecl}{\relax}
\BIBdecl

\bibitem{kawasaki2015}
Y.~Kawasaki, H.~Uga, S.~Kagiwada, and H.~Iyatomi, ``Basic study of automated
  diagnosis of viral plant diseases using convolutional neural networks,''
  \emph{International Symposium on Visual Computing}, pp. 638--645, 2015.

\bibitem{mohanty2016}
S.~P. Mohanty, D.~P. Hughes, and M.~Salath{\'e}, ``Using deep learning for
  image-based plant disease detection,'' \emph{Frontiers in Plant Science},
  vol.~7, p. 1419, 2016.

\bibitem{fuentes2017robust}
A.~Fuentes, S.~Yoon, S.~C. Kim, and D.~S. Park, ``A robust deep-learning-based
  detector for real-time tomato plant diseases and pests recognition,''
  \emph{Sensors}, vol.~17, no.~9, p. 2022, 2017.

\bibitem{fujita2018practical}
E.~Fujita, H.~Uga, S.~Kagiwada, and H.~Iyatomi, ``A practical plant diagnosis
  system for field leaf images and feature visualization,'' \emph{International
  Journal of Engineering \& Technology}, vol.~7, no. 4.11, pp. 49--54, 2018.

\bibitem{ferentinos2018deep}
K.~P. Ferentinos, ``Deep learning models for plant disease detection and
  diagnosis,'' \emph{Computers and Electronics in Agriculture}, vol. 145, pp.
  311--318, 2018.

\bibitem{quan2018}
Q.~H. Cap, K.~Suwa, E.~Fujita, H.~Uga, S.~Kagiwada, and H.~Iyatomi, ``An
  end-to-end practical plant disease diagnosis system for wide-angle cucumber
  images,'' \emph{International Journal of Engineering \& Technology}, vol.~7,
  no. 4.11, pp. 106--111, 2018.

\bibitem{suwa2019comparable}
K.~Suwa, Q.~H. Cap, R.~Kotani, H.~Uga, S.~Kagiwada, and H.~Iyatomi, ``A
  comparable study: Intrinsic difficulties of practical plant diagnosis from
  wide-angle images,'' \emph{IEEE International Conference on Big Data}, pp.
  5195--5201, 2019.

\bibitem{sa2016deepfruits}
I.~Sa, Z.~Ge, F.~Dayoub, B.~Upcroft, T.~Perez, and C.~McCool, ``Deepfruits: A
  fruit detection system using deep neural networks,'' \emph{Sensors}, vol.~16,
  no.~8, p. 1222, 2016.

\bibitem{bresilla2019single}
K.~Bresilla, G.~D. Perulli, A.~Boini, B.~Morandi, L.~Corelli~Grappadelli, and
  L.~Manfrini, ``Single-shot convolution neural networks for real-time fruit
  detection within the tree,'' \emph{Frontiers in Plant Science}, vol.~10, p.
  611, 2019.

\bibitem{tian2019apple}
Y.~Tian, G.~Yang, Z.~Wang, H.~Wang, E.~Li, and Z.~Liang, ``Apple detection
  during different growth stages in orchards using the improved yolo-v3
  model,'' \emph{Computers and Electronics in Agriculture}, vol. 157, pp.
  417--426, 2019.

\bibitem{deng2009imagenet}
J.~Deng, W.~Dong, R.~Socher, L.~J. Li, K.~Li, and L.~Fei-Fei, ``Imagenet: A
  large-scale hierarchical image database,'' \emph{IEEE Conference on Computer
  Vision and Pattern Recognition}, pp. 248--255, 2009.

\bibitem{krizhevsky2012imagenet}
A.~Krizhevsky, I.~Sutskever, and G.~E. Hinton, ``Imagenet classification with
  deep convolutional neural networks,'' \emph{Advances in Neural Information
  Processing Systems}, pp. 1097--1105, 2012.

\bibitem{real2019regularized}
E.~Real, A.~Aggarwal, Y.~Huang, and Q.~V. Le, ``Regularized evolution for image
  classifier architecture search,'' \emph{AAAI Conference on Artificial
  Intelligence}, vol.~33, pp. 4780--4789, 2019.

\bibitem{huang2019gpipe}
Y.~Huang, Y.~Cheng, A.~Bapna, O.~Firat, D.~Chen, M.~Chen, H.~Lee, J.~Ngiam,
  Q.~V. Le, Y.~Wu, and Z.~Chen, ``Gpipe: Efficient training of giant neural
  networks using pipeline parallelism,'' \emph{Advances in Neural Information
  Processing Systems}, pp. 103--112, 2019.

\bibitem{tan2019efficientnet}
M.~Tan and Q.~V. Le, ``Efficientnet: Rethinking model scaling for convolutional
  neural networks,'' \emph{International Conference on Machine Learning}, pp.
  6105--6114, 2019.

\bibitem{tom1995reconstruction}
B.~C. Tom and A.~K. Katsaggelos, ``Reconstruction of a high-resolution image by
  simultaneous registration, restoration, and interpolation of low-resolution
  images,'' \emph{International Conference on Image Processing}, vol.~2, pp.
  539--542, 1995.

\bibitem{schultz1996extraction}
R.~R. Schultz and R.~L. Stevenson, ``Extraction of high-resolution frames from
  video sequences,'' \emph{IEEE Transactions on Image Processing}, vol.~5,
  no.~6, pp. 996--1011, 1996.

\bibitem{patti2001artifact}
A.~J. Patti and Y.~Altunbasak, ``Artifact reduction for set theoretic super
  resolution image reconstruction with edge adaptive constraints and
  higher-order interpolants,'' \emph{IEEE Transactions on Image Processing},
  vol.~10, no.~1, pp. 179--186, 2001.

\bibitem{hirao2015prototype}
D.~Hirao and H.~Iyatomi, ``Prototype of super-resolution camera array system,''
  \emph{International Symposium on Visual Computing}, pp. 911--920, 2015.

\bibitem{quevedo2017underwater}
E.~Quevedo, E.~Delory, G.~Callic{\'o}, F.~Tobajas, and R.~Sarmiento,
  ``Underwater video enhancement using multi-camera super-resolution,''
  \emph{Optics Communications}, vol. 404, pp. 94--102, 2017.

\bibitem{freeman2002example}
W.~T. Freeman, T.~R. Jones, and E.~C. Pasztor, ``Example-based
  super-resolution,'' \emph{IEEE Computer Graphics and Applications}, vol.~22,
  no.~2, pp. 56--65, 2002.

\bibitem{komatsu2010super}
T.~Komatsu, Y.~Ueda, and T.~Saito, ``Super-resolution decoding of
  jpeg-compressed image data with the shrinkage in the redundant dct domain,''
  \emph{Picture Coding Symposium}, pp. 114--117, 2010.

\bibitem{dong2015image}
C.~Dong, C.~C. Loy, K.~He, and X.~Tang, ``Image super-resolution using deep
  convolutional networks,'' \emph{IEEE Transactions on Pattern Analysis and
  Machine Intelligence}, vol.~38, no.~2, pp. 295--307, 2015.

\bibitem{ledig2017photo}
C.~Ledig, L.~Theis, F.~Husz{\'a}r, J.~Caballero, A.~Cunningham, A.~Acosta,
  A.~Aitken, A.~Tejani, J.~Totz, Z.~Wang, and W.~Shi, ``Photo-realistic single
  image super-resolution using a generative adversarial network,'' \emph{IEEE
  Conference on Computer vision and Pattern Recognition}, pp. 4681--4690, 2017.

\bibitem{wang2018esrgan}
X.~Wang, K.~Yu, S.~Wu, J.~Gu, Y.~Liu, C.~Dong, Y.~Qiao, and C.~Change~Loy,
  ``Esrgan: Enhanced super-resolution generative adversarial networks,''
  \emph{European Conference on Computer Vision}, pp. 1--16, 2018.

\bibitem{goodfellow2014generative}
I.~Goodfellow, J.~Pouget-Abadie, M.~Mirza, B.~Xu, D.~Warde-Farley, S.~Ozair,
  A.~Courville, and Y.~Bengio, ``Generative adversarial nets,'' \emph{Advances
  in Neural Information Processing Systems}, pp. 2672--2680, 2014.

\bibitem{jolicoeur2018relativistic}
A.~Jolicoeur-Martineau, ``The relativistic discriminator: a key element missing
  from standard gan,'' \emph{International Conference on Learning
  Representations}, pp. 1--26, 2019.

\bibitem{kasturiwala2015adaptive}
S.~B. Kasturiwala and S.~Aladhake, ``Adaptive image superresolution for
  agrobased application,'' \emph{International Conference on Industrial
  Instrumentation and Control}, pp. 650--655, 2015.

\bibitem{yamamoto2017super}
K.~Yamamoto, T.~Togami, and N.~Yamaguchi, ``Super-resolution of plant disease
  images for the acceleration of image-based phenotyping and vigor diagnosis in
  agriculture,'' \emph{Sensors}, vol.~17, no.~11, p. 2557, 2017.

\bibitem{cap2019super}
Q.~H. Cap, H.~Tani, H.~Uga, S.~Kagiwada, and H.~Iyatomi, ``Super-resolution for
  practical automated plant disease diagnosis system,'' \emph{Annual Conference
  on Information Sciences and Systems}, pp. 1--6, 2019.

\bibitem{dai2020crop}
Q.~Dai, X.~Cheng, Y.~Qiao, and Y.~Zhang, ``Crop leaf disease image
  super-resolution and identification with dual attention and topology fusion
  generative adversarial network,'' \emph{IEEE Access}, vol.~8, pp.
  55\,724--55\,735, 2020.

\bibitem{giachetti2011real}
A.~Giachetti and N.~Asuni, ``Real-time artifact-free image upscaling,''
  \emph{IEEE Transactions on Image Processing}, vol.~20, no.~10, pp.
  2760--2768, 2011.

\bibitem{hughes2015open}
D.~Hughes and M.~Salath{\'e}, ``An open access repository of images on plant
  health to enable the development of mobile disease diagnostics,'' \emph{CoRR,
  abs/1511.08060}, 2015.

\bibitem{johnson2016perceptual}
J.~Johnson, A.~Alahi, and L.~Fei-Fei, ``Perceptual losses for real-time style
  transfer and super-resolution,'' \emph{European Conference on Computer
  Vision}, pp. 694--711, 2016.

\bibitem{saikawa2019aop}
T.~Saikawa, Q.~H. Cap, S.~Kagiwada, H.~Uga, and H.~Iyatomi, ``Aop: An
  anti-overfitting pretreatment for practical image-based plant diagnosis,''
  \emph{IEEE International Conference on Big Data}, pp. 5177--5182, 2019.

\bibitem{heusel2017gans}
M.~Heusel, H.~Ramsauer, T.~Unterthiner, B.~Nessler, and S.~Hochreiter, ``Gans
  trained by a two time-scale update rule converge to a local nash
  equilibrium,'' \emph{Advances in Neural Information Processing Systems}, pp.
  6626--6637, 2017.

\bibitem{maas2013rectifier}
A.~L. Maas, A.~Y. Hannun, and A.~Y. Ng, ``Rectifier nonlinearities improve
  neural network acoustic models,'' \emph{International Conference on Machine
  Learning}, vol.~30, no.~1, p.~3, 2013.

\bibitem{ioffe2015batch}
S.~Ioffe and C.~Szegedy, ``Batch normalization: Accelerating deep network
  training by reducing internal covariate shift,'' \emph{International
  Conference on Machine Learning}, pp. 448--456, 2015.

\bibitem{Simonyan15}
K.~Simonyan and A.~Zisserman, ``Very deep convolutional networks for
  large-scale image recognition,'' \emph{International Conference on Learning
  Representations}, pp. 1--14, 2015.

\bibitem{hiroki2018diagnosis}
T.~Hiroki, R.~Kotani, S.~Kagiwada, U.~Hiroyuki, and H.~Iyatomi, ``Diagnosis of
  multiple cucumber infections with convolutional neural networks,''
  \emph{Applied Imagery Pattern Recognition Workshop}, pp. 1--4, 2018.

\bibitem{toderici2017full}
G.~Toderici, D.~Vincent, N.~Johnston, S.~Jin~Hwang, D.~Minnen, J.~Shor, and
  M.~Covell, ``Full resolution image compression with recurrent neural
  networks,'' \emph{IEEE Conference on Computer Vision and Pattern
  Recognition}, pp. 5306--5314, 2017.

\bibitem{zhang2018unreasonable}
R.~Zhang, P.~Isola, A.~A. Efros, E.~Shechtman, and O.~Wang, ``The unreasonable
  effectiveness of deep features as a perceptual metric,'' \emph{IEEE
  Conference on Computer Vision and Pattern Recognition}, pp. 586--595, 2018.

\bibitem{kingma2015adam}
D.~P. Kingma and J.~Ba, ``Adam: A method for stochastic optimization,''
  \emph{International Conference on Learning Representations}, pp. 1--15, 2015.

\bibitem{he2016deep}
K.~He, X.~Zhang, S.~Ren, and J.~Sun, ``Deep residual learning for image
  recognition,'' \emph{IEEE Conference on Computer Vision and Pattern
  Recognition}, pp. 770--778, 2016.

\bibitem{cui2019class}
Y.~Cui, M.~Jia, T.~Y. Lin, Y.~Song, and S.~Belongie, ``Class-balanced loss
  based on effective number of samples,'' \emph{IEEE Conference on Computer
  Vision and Pattern Recognition}, pp. 9268--9277, 2019.

\end{thebibliography}
}

\end{document}